# A Novel Super Resolution Reconstruction of Low reoslution Images Progressively using DCT and Zonal Filter based Denoising

Liyakathunisa[1] and C.N .Ravi Kumar[2]

[1]Research Scholar, [2] Professor  & Head
Dept of Computer Science & Engineering,
Image Processing and Vision Lab
Sri Jayachamrajendra College of Engineering, Mysore, India.
[1]liyakath@indiatimes.com , [2]kumarcnr@yahoo.com

## Abstract

*Due to the factors like processing power limitations and channel capabilities images are often down sampled and transmitted at low bit rates resulting in a low resolution compressed image. High resolution images can be reconstructed from several blurred, noisy and down sampled low resolution images using a computational process know as super resolution reconstruction. Super-resolution is the process of combining multiple aliased low-quality images to produce a high resolution, high-quality image. The problem of recovering a high resolution image progressively from a sequence of low resolution compressed images is considered. In this paper we propose a novel DCT based progressive image display algorithm by stressing on the encoding and decoding process. At the encoder we consider a set of low resolution images which are corrupted by additive white Gaussian noise and motion blur. The low resolution images are compressed using 8 by 8 blocks DCT and noise is filtered using our proposed novel zonal filter. Multiframe fusion is performed in order to obtain a single noise free image. At the decoder the image is reconstructed progressively by transmitting the coarser image first followed by the detail image. And finally a super resolution image is reconstructed by applying our proposed novel adaptive interpolation technique. We have performed both objective and subjective analysis of the reconstructed image, and the resultant image has better super resolution factor, and a higher ISNR and PSNR. A comparative study done with Iterative Back Projection (IBP) and Projection on to Convex Sets (POCS),Papoulis Grechberg, FFT based  Super resolution Reconstruction shows that our method has out performed the previous contributions.*

## 1. Introduction

Super Resolution Reconstruction (SRR) is a process of producing a high spatial resolution image from one or more Low Resolution (LR) observation. It includes an alias free up sampling of the image thereby increasing the maximum spatial frequency and removing the degradations that arises during the image capture, Viz Blur and noise. An image is often corrupted by noise during acquisition and transmission. For instance in acquiring images with a CCD camera, light levels and sensor temperature are the major factors affecting the amount of noise in the resulting image. Images are also corrupted during transmission due to interference in the channel [3] [6]. Super resolution imaging has proved to be useful in many practical cases where multiple frames of the same scene are obtained. Some of the applications of super resolution imaging are in Astronomical imaging where it is possible to obtain different looks of the same scene. In satellite imagery when there is need for higher resolution images, then super-resolution imaging is a good choice. Medical imaging is a very important application area for image super-resolution. Many medical types of equipment as the Computer Aided Tomography (CAT), the Magnetic Resonance Images (MRI), or the Echography or Mammography images allow the acquisition of several images, which can be combined in order to obtain a higher-resolution image. CCD size limitations and shot noise prevents obtaining very high resolution images directly. Super-resolution reconstruction can be used to get desired high resolution image [2][3].

In recent years several research groups have started to address the goal of resolution augmentation in medical imaging as software post processing challenge. The motivation for this initiative emerged following major advances in the domains of image and video processing that indicated the possibility of enhancing the resolution using "Super Resolution" algorithms. Super Resolution deals with the task of using several low resolution images from a particular imaging system to estimate, or reconstruct, the high resolution image.

The emergence of new easy, fast and reliable techniques of image storage and Internet transmission has stirred up the practice of medicine. For example, patients could get immediate diagnosis of any specialist located anywhere [10].
In classical image encoding systems, the last compression operation consists in entropy coding, generally based on variable length codes. If the coding process produces a single bit stream, as soon as one bit is lost during the transmission of encoded data, the whole image is then lost. For this reason, progressive transmission of information constitutes a key issue in the domain of telemedicine and teleastronomy [10].

Progressive image transmission is a method which allows to obtain a high quality version of the original image from the minimal amount of data. In "progressive" mode of transmission, namely as more bits are transmitted, better quality reconstructed images can be produced at the receiver. The receiver need not wait for all of the bits to arrive before decoding the image; in fact, the decoder can use each additional received bit to improve somewhat upon the previously reconstructed image. In Progressive Image Transmission (PIT), an approximate image is built up quickly in one stage and refined progressively in later. Some of the advantages of a progressive image transmission are that it allows to interrupt the transmission when the quality of the received image has reached a desired accuracy or when the receiver recognizes that the image is not interesting or only needs a specific portion of the complete image or images.

In this paper we propose a novel DCT based approach with efficient denoising, which progressively reconstructs a high resolution image from a set of low resolution images. Our algorithm explores applications such as Astronomical images (Land sat 7) and Medical Images (Mammograms).

The flow of the topics is as follows, In section II Mathematical Formulation for Super Resolution Model is created and described, in section III existing super resolution reconstruction techniques have been discussed, section IV presents the proposed method for SRR of LR images, section V The proposed progressive transmission for super resolution reconstruction is discussed, section VI provides the proposed algorithm for Super Resolution Reconstruction of LR images. Experimental results are presented and discussed in section VII and finally section VIII consists of conclusion.

## 2. MATHEMATICAL FORMULATION FOR THE SUPER RESOLUTION MODEL.

In this section we give the mathematical model for super resolution image reconstruction from a set of Low Resolution (LR) images.

Let us consider the low resolution sensor plane by $M_1$ by $M_2$. The low resolution intensity values are denoted as $\{y(i,j)\}$ where $i = 0...M_1 -1$ and $j = 0...M_2 -1$; if the down sampling parameters are $q_1$ and $q_2$ in horizontal and vertical directions, then the high resolution image will be of size $q_1 M_1 \times q_2 M_2$. We assume that $q_1 = q_2 = q$ and therefore the desired high resolution image Z will have intensity values $\{z(k,l)\}$ where $k = 0...qM_1 - 1 \text{ and } l = 0...qM_2 - 1$.

Given $\{z(k,l)\}$ the process of obtaining down sampled LR aliased image $\{y(i,j)\}$ is

$$y(i,j) = \frac{1}{q^2} \sum_{k=qi}^{(q+1)i-1} \sum_{l=qj}^{(q+1)j-1} z(k,l) \tag{1}$$

i.e. the low resolution intensity is the average of high resolution intensities over a neighborhood of $q^2$ pixels.

We formally state the problem by casting it in a Low Resolution restoration frame work. There are P observed images $\{Y_m\}_{m=1}^P$ each of size $M_1 \times M_2$ which are decimated, blurred and noisy versions of a single high resolution image Z of size $N_1 \times N_2$ where $N_1 = qM_1$ and $N_2 = qM_2$. After incorporating the blur matrix, and noise vector, the image formation model is written as

$$Y_m = H_m DZ + \eta_m \text{ Where m=1...P} \tag{2}$$

Here D is the decimation matrix of size $M_1 M_2 \times q^2 M_1 M_2$, H is Blurring function or the Point spread function (PSF) of size $M_1 M_2 \times M_1 M_2$, $\eta_m$ is $M_1 M_2 \times 1$ noise vector and P is the number of low resolution observations Stacking P vector equations from different low resolution images into a single matrix vector

$$\begin{bmatrix} y_1 \\ . \\ . \\ . \\ y_p \end{bmatrix} = \begin{bmatrix} DH_1 \\ . \\ . \\ DH_p \end{bmatrix} Z + \begin{bmatrix} \eta_1 \\ . \\ . \\ \eta_p \end{bmatrix} \tag{3}$$

The matrix D represents filtering and down sampling process of dimensions $q^2 M_1 M_2 \times 1$ where q is the resolution enhancement factor in both directions. Under separability assumptions, the matrix D which transforms the $qM_1 \times qM_2$ high resolution image to $N_1 \times N_2$ low resolution images where $N_1 = qM_1$, $N_2 = qM_2$ is given by

$$D = D_1 \otimes D_1 \tag{4}$$

Where $\otimes$ represents the kronecker product, and the matrix $D_1$ represents the one dimensional low pass filtering and down sampling. When q=2 the matrix $D_1$ will be given by

$$D_1 = \frac{1}{2} \begin{bmatrix} 11\ 00\ 00\ .....00 \\ 00\ 11\ 00......00 \\ : \quad : \quad : \quad\quad : \\ : \quad : \quad : \quad\quad : \\ 00\ 00 \quad\quad\quad 11 \end{bmatrix} \text{ and } D = \frac{1}{2^2} \begin{bmatrix} 11\ 00\ 00\ .....00 \\ 00\ 11\ 00......00 \\ : \quad : \quad : \quad\quad : \\ : \quad : \quad : \quad\quad : \\ 00\ 00 \quad\quad\quad 11 \end{bmatrix} \tag{5}$$

Typically the blur is described mathematically with a point spread function (PSF); a function that specifies how the points in the image are distorted, blur can arrive from a variety of sources, such as atmospheric turbulence, out of focus, and motion blur. The blur can be classified as either spatially invariant or spatially variant, and we need to know what kind of boundary conditions is to be used.

Spatially invariant implies that the blur is independent of position. That is the blurred object will look the same regardless of its position in the image. Spatially variant implies that the blur depends on position. That is an object in an observed image may look different if its position is changed. If we assume that the blur is spatially invariant then the PSF is represented by the image of a single point source. In this case, the structure of H depends on the boundary condition. Images are shown only in a finite region, but points near the boundary of a blurred image are likely to have been affected by information outside the field of view. Since this information is not available, for computational purposes, we need to make some assumption about the boundary conditions. Periodic boundary conditions imply that the image repeats itself endlessly in all directions; periodic boundary conditions imply that H is a block circulant matrix with circulant blocks (BCCB).

The square matrix H of dimensions $PN_1 \times PN_2$ represents intra channel and inter channel blur operators. i.e. 2D convolution of channel with shift –invariant blurs. The blur matrix is of the form

$$H_I = \begin{bmatrix} H_{(0)} & H_{(1)} & \cdots & H_{(M-1)} \\ H_{(M-1)} & H_{(0)} & \cdots & H_{(M-2)} \\ \vdots & \vdots & \cdots & \vdots \\ \vdots & \vdots & \cdots & \vdots \\ H_{(1)} & H_{(2)} & \cdots & H_{(0)} \end{bmatrix} \quad (6)$$

and it is circulant at the block level. In general each $H_{(i)}$ is an arbitrary $PM_1 \times PM_2$, but if shift invariant circular convolution is assumed H(i) becomes

$$H_{(i)} = \begin{bmatrix} H_{(i,0)} & H_{(i,1)} & \cdots & H_{(i,M-1)} \\ H_{(i,M-1)} & H_{(i,0)} & \cdots & H_{(i,M-2)} \\ \vdots & \vdots & \cdots & \vdots \\ \vdots & \vdots & \cdots & \vdots \\ H_{(i,1)} & H_{(i,2)} & \cdots & H_{(i,0)} \end{bmatrix} \quad (7)$$

which is also circulant at the block level $H_{(i,j)}$. Each $P \times P$ sub matrix (sub blocks) has the form.

$$H_{(i,j)} = \begin{bmatrix} H_{11(i,j)} & H_{12(i,j)} & \cdots & H_{1p(i,j)} \\ H_{21(i,j)} & H_{22(i,j)} & \cdots & H_{2p(i,j)} \\ \vdots & \vdots & \cdots & \vdots \\ \vdots & \vdots & \cdots & \vdots \\ H_{p1(i,j)} & H_{p2(i,j)} & \cdots & H_{pp(i,j)} \end{bmatrix} \quad (8)$$

Where $H_{ii(m)}$ is intra channel blurring operator, $H_{ij(m)} i \neq j$ is an inter channel blur i.e. P x P non circulant blocks are arranged in a circulant fashion, it's called Block Semi-Block Circulant (BSBC) which can be easily solved using FFT based techniques.

## 3. EXISTING SUPER RESOLUTION RECONSTRUCTION

Super Resolution algorithm can be classified as Reconstruction based methods and Learning based methods. In reconstruction based super resolution, multiple low resolution observations of the same scene is required to estimate the high resolution counterpart. In learning based methods it is proposed that only one low resolution image includes adequate information to predict the details to super resolve the image. Learning based super resolution is a popular technique that uses application dependent priors to infer the missing details in low resolution images. Learning based super resolution algorithms extract a relationship between high resolution images and their corresponding low resolution ones which are used as training data.

Numerous Super Resolution algorithms have been proposed in literature [2],[5],[18],[22],[23],[24],[28]. Dating back to the frequency domain approach of Huang and Tsai [28] 1984 to [38] in 2010. Although a detail survey is provided by Borman [2] and Park [5], it is felt that there is need to discuss super resolution techniques used before 2000 and after 2000 both on learning based and reconstruction based.

Frequency Domain Techniques [22] [23] [28]; Tsai and Huang [28] in 1984, proposed the multiframe super resolution problem. Motivated by the need for improved resolution images from Landsat image data. They assume a purely translational motion and solve the dual problem of registration and restoration. Super Resolution Reconstruction by Vandewalle et al [22] and Deepesh Jain [23] focuses on accurate image registration necessary for a perfect super resolution reconstruction. Their registration part is based on estimating the shift and rotation between the input image and the reference image using Fourier Transforms, and aligning them. A Blind Super Resolution (BSR) reconstruction was proposed by F. Sroubek, J.Flusser [24] and Tao Hongjiu [35]. The problem of recovering a high resolution image from a sequence of low resolution DCT based compressed observation was proposed by S.C.Park and Katsaggeloes [36] in 2004. In recent years lot of research is being carried out in Learning based super resolution [25][37]. It consists of two basic processes namely a training process and a super resolution or synthesis process. Baker and Kanade[26] presented a pioneering work on hallucinating face image based on Bayesian formulation. Freeman et al [22] posed the image super resolution as the problem of estimating missing high frequency details by interpolating input low resolution image into the desired scale. Learning based super resolution using multi resolution wavelet approach was proposed by Kim and Hwang et al [37]. A robust Super resolution reconstruction with efficient denoising using wavelets is proposed in 2010 [38].

## 4. PROPOSED METHOD FOR SRR OF LR IMAGES

Our Low Resolution mammogram images consist of the degradations such as geometric registration wrap, sub sampling, blurring and additive noise. Based on these phenomena our implementation consists of the following phases.

- Image registration

- Restoration of the registered low resolution images using our proposed zonal filter based denoising.

- Interpolation of the restored image to obtain a super resolution image.

## 4.1. Image Registration

In Super Resolution Reconstruction the first pre-processing task of utmost importance is accurate registration of the acquired images. It is a process of overlaying two or more images of the same scene taken at different times, from different view points and or by different sensors. Typically one image called the base image is considered the reference, to which the other images called input images are compared. The objective is to bring the input image in alignment with the base image by applying a spatial transformation to the input image. Spatial transformations maps locations in one image into a new location in another image. Image registration is an inverse problem as it tries to estimate from sampled images $Y_m$, the transformation that occurred between the views $Z_m$ considering the observation model of Eq(2). It is also dependent on the properties of the camera used for image acquisition like sampling rate (or resolution) of sensor, the imperfection of the lens that adds blur, and the noise of the device. As the resolution decreases, the local two dimensional structure of an image degrades and an exact registration of two low resolution images becomes increasingly difficult. Super resolution reconstruction requires a registration of high quality. The registration technique considered in our research is based on Fast Fourier Transform proposed by Fourier Mellin and DeCastro and Morandi [8][9]. The transformation considered in our research is rotation, translation and shift estimation. Let us consider the translation estimation, the Fourier transform of the function is denoted by $F\{f(x,y)\}$ or $\hat{f}(w_x, w_y)$. The shift property of the Fourier transform is given by

$$F\{f(x+\Delta x, y+\Delta y)\} = \hat{f}(w_x, w_y) e^{i(w_x \Delta x + w_y \Delta y)} \quad (9)$$

Eq.9 is the basis of the Fourier based translation estimation algorithms. Let $I_1(x, y)$ be the reference image and $I_2(x, y)$ is the translated version of the base image, i.e.

$$I_1(x, y) = I_2(x+\Delta x, y+\Delta y) \quad (10)$$

By applying the Fourier transform on both the sides of Eq. (10). We get

$$\hat{I}_1(w_x, w_y) = \hat{I}_2(w_x, w_y) e^{i(w_x \Delta x + w_y \Delta y)} \quad (11)$$

or equivalently,

$$\frac{\hat{I}_1(w_x, w_y)}{\hat{I}_2(w_x, w_y)} = e^{i(w_x \Delta x + w_y \Delta y)} \quad (12)$$

$$corr(x, y) \cong F^{-1}\left(\frac{\hat{I}_1(w_x, w_y)}{\hat{I}_2(w_x, w_y)}\right) = \delta(x+\Delta x, y+\Delta y) \quad (13)$$

For discrete images we replace the Fourier Transform in the computation above with Fast Fourier Transform, and $\delta(x+\Delta x, y+\Delta y)$ is replaced by a function that has dominant maximum at $(\Delta x, \Delta y)$ as

$$(\Delta x, \Delta y) = \arg\max\{corr(x, y)\} \quad (14)$$

Calculate the cross power spectrum by taking the complex conjugate of the second result. Multiplying the Fourier Transform together element wise, and normalizing this product element wise.

$$corr(w_x, w_y) \cong \frac{\hat{I}_1(w_x, w_y)}{\hat{I}_2(w_x, w_y)} \bullet \left| \frac{\hat{I}_1(w_x, w_y)}{\hat{I}_2(w_x, w_y)} \right| \quad (15)$$

$$corr(w_x, w_y) = R \cong \frac{\hat{I}_1(w_x, w_y)\hat{I}_2^*(w_x, w_y)}{\left|\hat{I}_2(w_x, w_y)\right|\left|\hat{I}_1^*(w_x, w_y)\right|} = e^{i(w_x \Delta x + w_y \Delta y)} \quad (16)$$

where * denotes the complex conjugate. The normalized cross correlation is obtained by applying the inverse Fourier transform. i.e. $r = F^{-1}\{R\}$, determine the location of the peak in r. This location of the peak is exactly the displacement needed to register the images.

$$(\Delta x, \Delta y) = \arg\max\{r\} \quad (17)$$

The angle of rotation is estimated by converting the Cartesian coordinates to log polar form. We observe that the sum of a cosine wave and a sine wave of the same frequency is equal to phase shifted cosine wave of the same frequency. That is if a function is written in Cartesian form as

$$v(t) = A\cos(t) + B\sin(t) \quad (18)$$

Then it may also be written in polar form as

$$v(t) = c\cos(t - \varphi) \quad (19)$$

We may write the Eq (19) in polar form as

$$Y = y(x) = \frac{a_0}{2} + \sum_{k=1}^{N} m_k \cos(2\pi f_k x - \varphi_k) \quad (20)$$

Where

$$m_k = \sqrt{a_k^2 + b_k^2} \ldots (magnitude)$$
$$\varphi_k = \tan^{-1}\left(\frac{b_k}{a_k}\right) \ldots (Phase) \quad (21)$$

The Shift is estimated by finding cross power spectrum and computing Eq.(16). We obtain the normalized cross correlation by applying the inverse Fourier Transform. i.e. $r = F^{-1}\{R_2\}$, determines the location of the peak in r. This location of the peak is exactly the shift $I(x_0, y_0)$ needed to register the images. Once we estimate the angle of rotation and translation and shift a new image is constructed by reversing the angle of rotation, translation and shift.

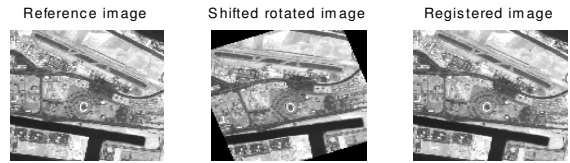

Fig 3: Registered Image

## 4.2. Proposed Blind Super Resolution Restoration using DCT

Images are obtained in areas ranging from everyday photography, astronomy, remote sensing, medical imaging, microscopy and many more. In each of mere cases there is an underlying object or scene we wish to observe. The original image or the true image is the ideal representation of the observed scene. Yet the observation process is never perfect, there is uncertainty in the measurement occurring as blur, noise and other degradations in the recorded images. Image restoration aims to recover an estimate of the original image from the degraded observations. Classical image restoration seeks an estimate of the true image assuming the blur is known, whereas blind image restoration tackles the much more difficult but realistic problem where the degradations are unknown.

The low resolution observation model of Eq. (2) is considered. We formally state by casting the problem in multi channel restoration format, the noise is AWGN and the blur is considered as between channels and within channel of the low resolution images.

### 4.2.1. Proposed Zonal filter based denosing

Transform domain features can be extracted by zonal filtering or zonal coding. Transform domain compression works by only sending part of the transform, if we apply a zonal mask to the transformed blocks and encode only the nonzero elements, then the method is called zonal coding or zonal filter. Various zonal mask suggested by A. K. Jain [11] are used in implementation. In transform domain coding only a small zone of transformed image is transmitted. Let $N_t$ be the number of transmitted samples. We define a zonal mask as the array

$$m(i, j) = \begin{cases} 1 & i,j \; \varepsilon \; I_t \\ 0 & \text{otherwise} \end{cases} \qquad (22)$$

which takes the unity value in the zone of largest $N_t$ variances of the transformed samples. A zonal mask is constructed by placing a 1 in the locations of maximum variance and a 0 in all other locations. Coefficients of maximum variance are located around the origin of an image transform, Fig.4. Shows the typical zonal mask [6][11]. In Zonal coding, transform coefficients are selected using a fix filter mask, which has been formed to fit the maximum variance of an average image, knowing which transform coefficients that are going to be retained makes it possible to optimize the transform computation, only the selected transform coefficients are need to be computed.

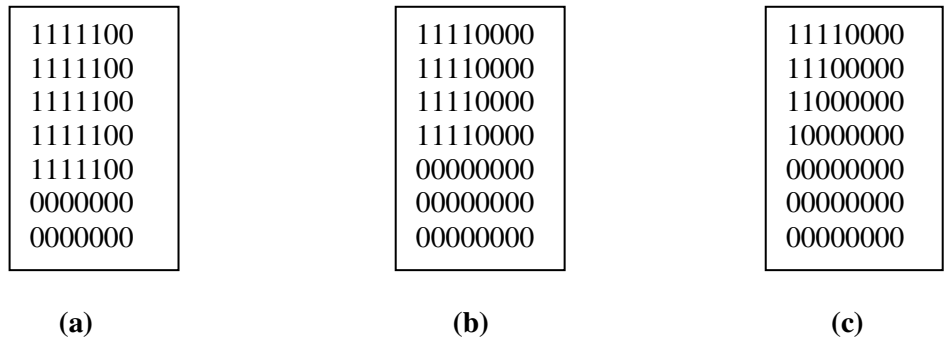

Fig 4: Different Zonal Masks

### 4.2.2 Image Deblurring

The blur is removed using the deblurring filter, i.e. by using Iterative Blind Deconvolution (IBD)[12]. There are numerous situations in which the point-spread function is not explicitly known, and the true image Z must be identified directly from the observed image Y by using partial or no information about the true image and the point-spread function. In these cases, we have more difficult problem of blind deconvolution. Blind deconvolution is a deconvolution technique that permits recovery of the target scene from a single or set of blurred images in the presence of a poorly determined or unknown PSF.

Blind deconvolution can be performed iteratively where each iteration improves the estimation of the PSF and the scene IBD starts with an initial estimate of the restored image, an initial estimate of the PSF restoring the image is by making an initial estimate of what the PSF and image are. One of the constraints that we apply to the image is that of finite support. Finite support basically says that the image does not exist beyond a certain boundary. The first set of Fourier constraints involve estimating the PSF using the FFT of the degraded image and FFT of the guessed PSF

$$H_k(u,v) = \frac{G(u,v) conj(\bar{F}(u,v))}{|\bar{F}(u,v)|^{^2} + alpha|F(u,v)|^{^2}} \quad (23)$$

By applying IFFT $H_k(u,v)$, we obtain the PSF. The true image is restored by deconvolution of the PSF with the degraded image. Hence the second set of constraints involve

$$F_k(u,v) = \frac{G(u,v) conj(\bar{H}(u,v))}{|\bar{H}(u,v)^{^2}| + alpha|\bar{H}(u,v)|^{^2}} \quad (24)$$

The blur constraints that are applied are from the assumptions that we know or have some knowledge of the size of the PSF.

### 4.3 Adaptive Interpolation for SRR

The algorithm works in four phases: In the first phase the wavelet based fused image is expanded. Suppose the size of the fused image is n x m.
The image will be expanded to size (2n-1) x (2m-1).

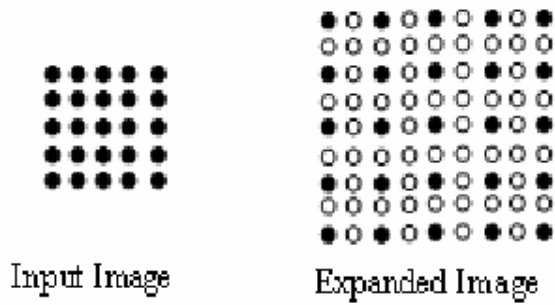

Fig 5: High Resolution Grid

In the Fig 5 solid circles show original pixels and hallow circles show undefined pixels. In the remaining three phases these undefined pixels will be filled.

The second phase of the algorithm is most important one. In this phase the interpolator assigns value to the undefined pixels

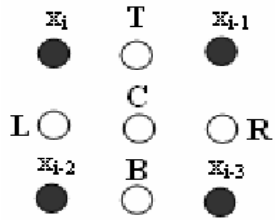

Fig 6: HR unit cell with undefined pixels Top, Center, Bottom, Left, Right dented by T,C,B,L,R.

The undefined pixels are filled by following mutual exclusive condition.

Uniformity: select the range $(X_i, X_{i-1}, X_{i-2}, X_{i-3})$ and a Threshold T.

$$\text{if range } (X_i, X_{i-1}, X_{i-2}, X_{i-3}) < T \text{ Then}$$
$$C = (X_i + X_{i-1} + X_{i-2} + X_{i-3})/4 \tag{25}$$

$$\begin{aligned}&\textit{if } \text{there is edge in NW-SE Then} \\ &C=(X_i + X_{i-1})/2 \\ &\textit{if } \text{there is edge in NS Then} \\ &T=(X_i + X_{i-1})/2 \text{ and } B =(X_{i-2} + X_{i-3})/2 \\ &\textit{if } \text{there is edge in EW Then} \\ &L=(X_i + X_{i-2})/2 \text{ and } R =(X_{i-1} + X_{i-3})/2\end{aligned} \tag{26}$$

In this phase, approximately 85% of the undefined pixels of HR image are filled.

In the third phase the algorithm scans magnified image, line by line and looks for those pixels which are left undefined in the previous phase.

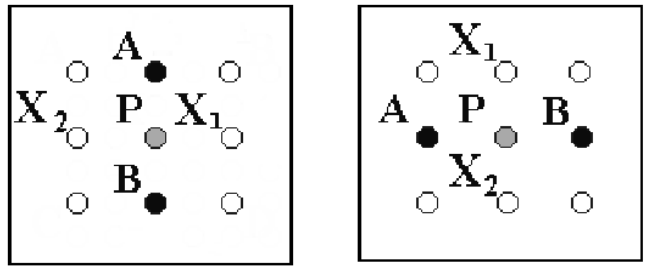

Fig 7: Layout referred in the phase 3

The algorithm checks for the layout as shown in Fig 7.

$$\begin{aligned} &\textit{if} \text{ there is edge } X_1 X_2 \text{ then } P = (X_1 + X_2)/2 \\ &\textit{else if} \text{ there is edge AB then } P = (A + B)/2 \end{aligned} \quad (27)$$

In the fourth phase all the undefined pixels will be filled. If there are any undefined pixels left then the median of the neigh boring pixels is calculated and assigned. We call our interpolation method adaptive as the interpolator selects and assigns the values for the undefined pixels based on mutual exclusive condition.

## 5. PROPOSED PROGRESSIVE TRANSMISSION FOR SUPER RESOLUTION RECONSTRUCTION.

The low resolution observation model of Eq. 2 is considered. In this paper we propose a novel progressive image compression algorithm for super resolution reconstruction using DCT. At the encoder the images have to be compressed before transmission. A 2D DCT is applied for each low resolution image whose size is chosen as 256x256. The low resolution images are divided into sub image of size 8x8, with a total of 32 non overlapping blocks. DCT is applied on each 8x8 blocks producing one DC coefficient and AC coefficients. In order to remove the noise which is present during acquisition, we encode each Block DCT image by applying the zonal filters to the low frequency components and discard the high frequency components, which results in denoised image. Once the images have been deblurred and denoised. The images are fused using maximum frequency fusion to obtain a single image. The zonal masks as shown in the Fig .4. (c). is used in our implementation which results in 1DC and 9 AC coefficients shown below in Fig. 8. (c).

| DC | AC1 | AC2 | AC3 | 0 | 0 | 0 | 0 |
|---|---|---|---|---|---|---|---|
| AC4 | AC5 | AC6 | AC7 | 0 | 0 | 0 | 0 |
| AC8 | AC9 | AC10 | AC11 | 0 | 0 | 0 | 0 |
| AC12 | AC13 | AC14 | AC15 | 0 | 0 | 0 | 0 |
| AC16 | AC17 | AC18 | AC19 | 0 | 0 | 0 | 0 |
| AC20 | AC21 | AC22 | AC23 | 0 | 0 | 0 | 0 |
| 0 | 0 | 0 | 0 | 0 | 0 | 0 | 0 |
| 0 | 0 | 0 | 0 | 0 | 0 | 0 | 0 |

(a)

| DC | AC1 | AC2 | AC3 | 0 | 0 | 0 | 0 |
|---|---|---|---|---|---|---|---|
| AC4 | AC5 | AC6 | AC7 | 0 | 0 | 0 | 0 |
| AC8 | AC9 | AC10 | AC11 | 0 | 0 | 0 | 0 |
| AC12 | AC13 | AC14 | AC15 | 0 | 0 | 0 | 0 |
| 0 | 0 | 0 | 0 | 0 | 0 | 0 | 0 |
| 0 | 0 | 0 | 0 | 0 | 0 | 0 | 0 |
| 0 | 0 | 0 | 0 | 0 | 0 | 0 | 0 |
| 0 | 0 | 0 | 0 | 0 | 0 | 0 | 0 |

(b)

| DC | AC1 | AC2 | AC3 | 0 | 0 | 0 | 0 |
|---|---|---|---|---|---|---|---|
| AC4 | AC5 | AC6 | 0 | 0 | 0 | 0 | 0 |
| AC7 | AC8 | 0 | 0 | 0 | 0 | 0 | 0 |
| AC9 | 0 | 0 | 0 | 0 | 0 | 0 | 0 |
| 0 | 0 | 0 | 0 | 0 | 0 | 0 | 0 |
| 0 | 0 | 0 | 0 | 0 | 0 | 0 | 0 |
| 0 | 0 | 0 | 0 | 0 | 0 | 0 | 0 |
| 0 | 0 | 0 | 0 | 0 | 0 | 0 | 0 |

(c)

Figure 8: Zonal Filtered coefficients

| **Stage1** DC | **Stage2** AC1 | **Stage3** AC2 | **Stage4** AC3 | **Sarge5** AC4 | **Satge6** AC5 | **Satge7** AC6 | **Satge8** AC7 | **Satge9** AC8 | **Satge10** AC9 |
|---|---|---|---|---|---|---|---|---|---|

Figure 9: Stages of Bit Transmission

The sender transmits these DCT coefficients of each stage individually. Then the receiver reconstructs the image based on the received coefficients. At the decoder in the first stage content received, perform inverse DCT on the DC coefficients and thus a coarse image is obtained. In the second stage the sender transmits the DC and AC1 coefficients. Also perform inverse DCT on the DC and AC1 coefficients the reconstructed quality is enhanced. The remaining stages are processed in the similar fashion. As soon as all the stages are received, a high resolution image is reconstructed. Then apply our adaptive interpolation discussed in section 3 to the reconstructed image to obtain an image with double, quadruple the resolution to that of the original image. The obtained Super resolution image performs better when compared to other traditional approaches.

## 6. PROPOSED ALGORITHM FOR SUPER RESOLUTION RECONSTRUCTION OF LOW RESOLUTION IMAGES

Our proposed novel super resolution consists of following consecutive steps:

**At the encoder**:

Step 1: Three input low resolution blurred, noisy, under sampled, rotated, shifted images are considered.

i.e. $I_1(i, j), I_2(i, j), I_3(i, j)$ where $i=1…N$, $j=1….N$

Step 2: The images are first preprocessed, i.e. registered using FFT based algorithm, as discussed in section 4.1.

Step 3: Each LR image is divided into 8x8 non overlapping blocks; DCT is applied to each block.

Step 4**:** Zonal mask is applied to low frequency components and high frequency components are discarded.

Step 5: Iterative Blind Deconvolution (IBD) is applied to remove the blur present in the images.

Step 6: Restoration is performed in order to remove the blur and noise present in the image.

Step 7: The restored images are fused using maximum frequency fusion.

**At the Decoder:**

Step 1: In the first stage of reconstruction Inverse DCT is applied to DC coefficients and then transmitted, a coarse image is reconstructed at the receiver.

Step 2: In the second stage of transmission Inverse DCT is applied to DC and the first AC coefficient in each 8 x 8 blocks, the reconstructed image quality is enhanced.

Step 3: The above procedure is repeated to the remaining AC coefficients, progressively at each stage the resolution is increased and the quality of the image is improved resulting in a high resolution image.

Step 4: Finally in order to obtain a super resolved image, an image with double the resolution as that of the original image our proposed interpolation technique is applied.

## 7. RESULTS AND DISCUSSIONS:

### Case 1: Abudhabi Stadium

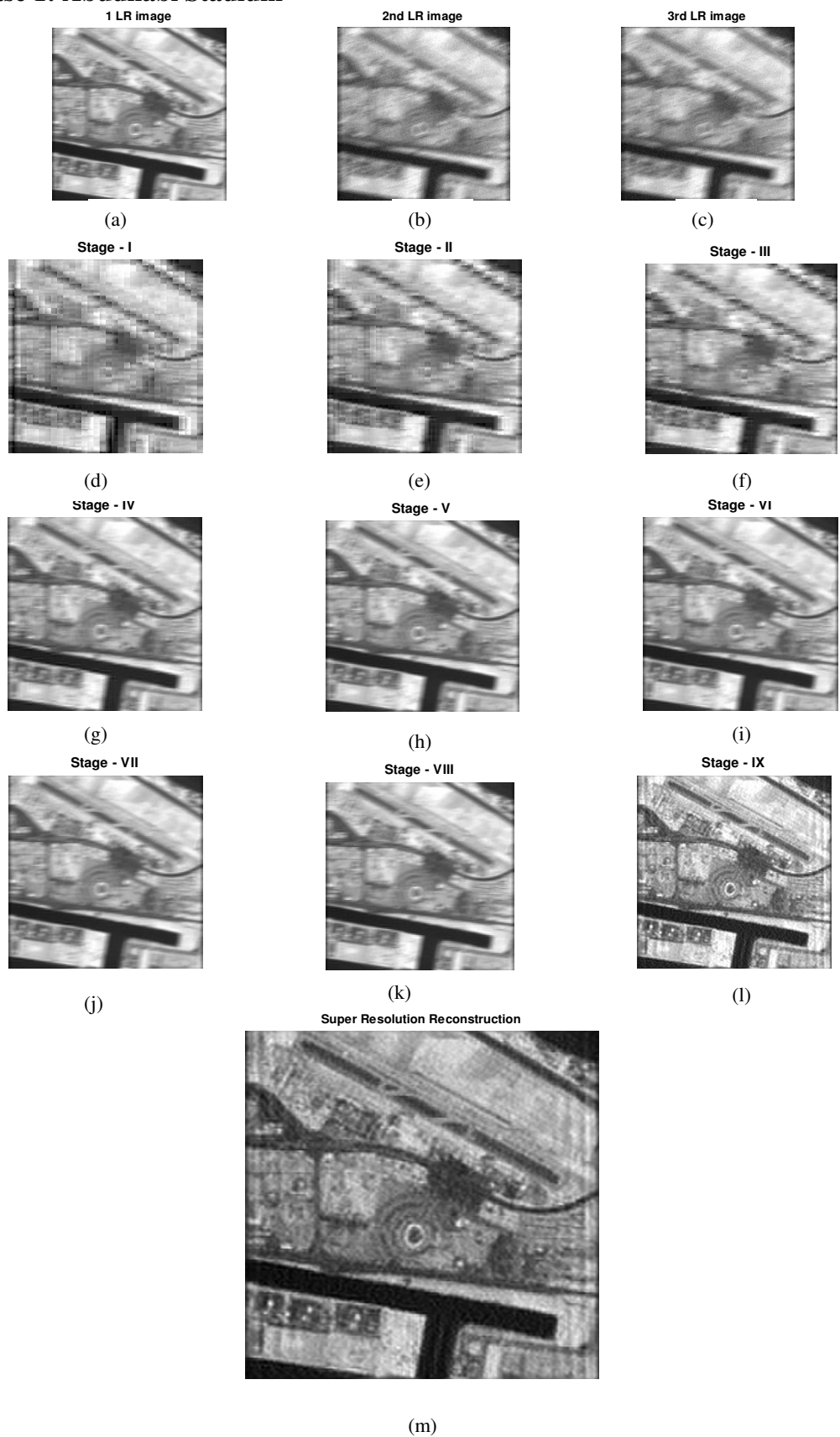

**FIGURE 10:** Three Low Resolution SAR Images are shown in (a)-(c), different stages of progressive reconstruction is shown in (d)-(l), where as (m) is SRR image

**Case 2: Mammogram**

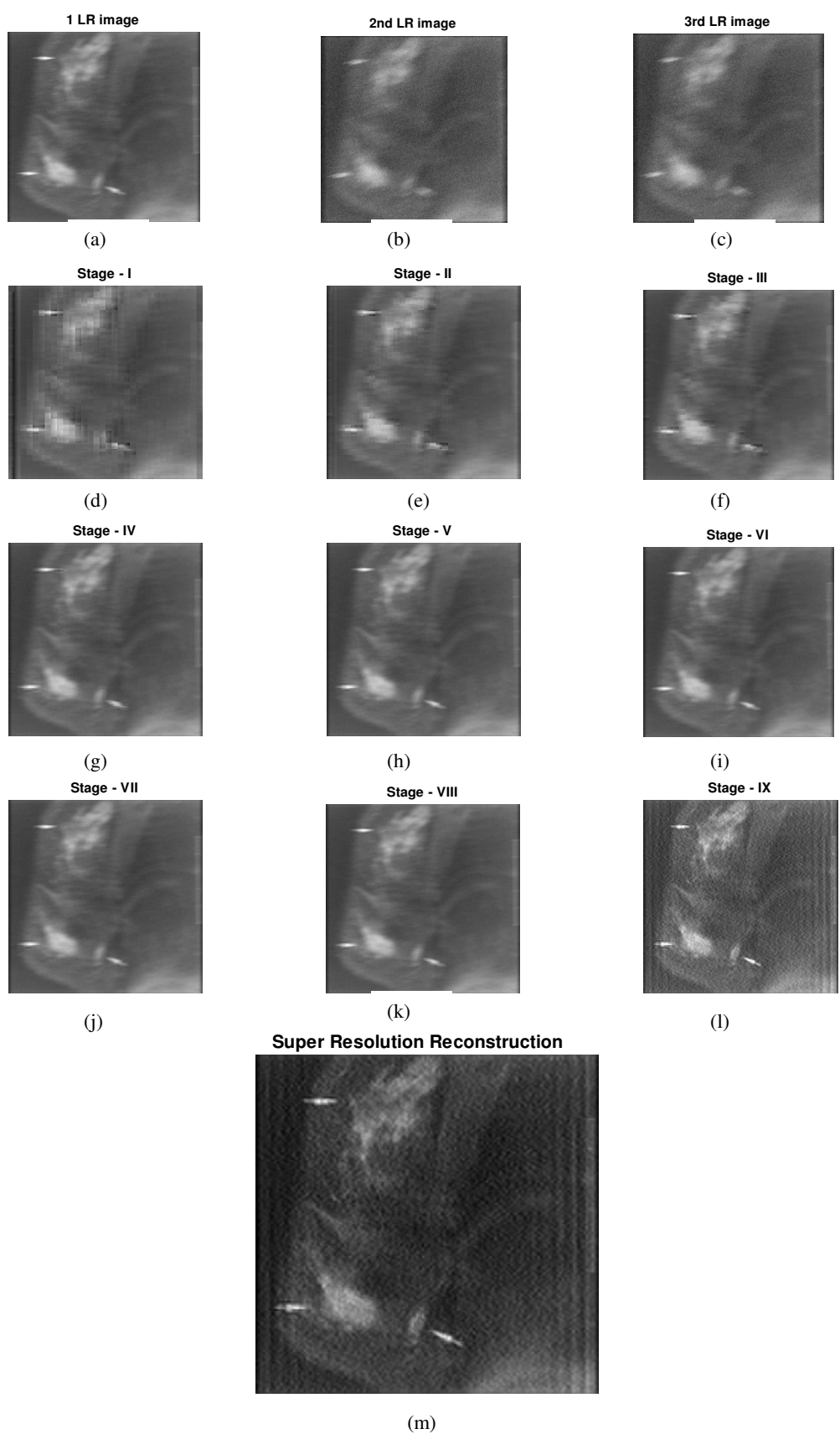

**FIGURE 11:** Three Low Resolution Mammogram Images are shown in (a)-(c), different stages of progressive reconstruction is shown in (d)-(l), where as (m) is SRR image

## 7.1 Performance Evaluation

The performance of the algorithm for various images at different blur and noise levels is studied and the results for two cases are shown in Fig.10. and Fig 11. The quantities for comparison are defined as follows, and Table I and II display the quantitative measures.

**1) Improvement in Signal-to-Noise Ratio (ISNR)**

For the purpose of objectively testing the performance of the restored image, Improvement in signal to noise ratio (ISNR) is used as the criteria which is defined by

$$\text{ISNR} = 10 \log_{10} \frac{\sum_{i,j}[f(i,j) - y(i,j)]^2}{\sum_{i,j}[f(i,j) - g(i,j)]^2} \quad (28)$$

Where j and i are the total number of pixels in the horizontal and vertical dimensions of the image; f(i, j), y(i, j) and g(i, j) are the original, degraded and the restored image.

**2) The MSE and PSNR of the reconstructed image is**

$$MSE = \frac{\sum [f(i,j) - F(I,J)]^2}{N^2} \quad (29)$$

Where f(i, j) is the source image F(I,J) is the reconstructed image, which contains N x N pixels

$$PSNR = 20 \log_{10}\left(\frac{255}{RMSE}\right) \quad (30)$$

**3) Super Resolution Factor**

$$SRF = \frac{\sum_{i=1}^{M}\sum_{j=1}^{N}(F(i,j) - f(i,j))^2}{\sum_{i=1}^{M}\sum_{j=1}^{N}(y(i,j) - f(i,j))^2} \quad (31)$$

**4) MSSIM**
The structural similarity (SSIM) index is defined in [29] by equations

$$SSIM(f,F) = \frac{(2\mu_f \mu_F + C_1)(2\sigma_F + C_2)}{(\mu_f^2 + \mu_F^2 + C_1)(\sigma_f^2 + \sigma_F^2 + C_2)} \quad (32)$$

$$MSSIM(f,F) = \frac{1}{G}\sum_{p=1}^{G} SSIM(f,F) \quad (33)$$

The Structural SIMilarity index between the original image and reconstructed image is given by SSIM, where $\mu_f$ and $\mu_F$ are mean intensities of original and reconstructed images, $\sigma_f$ and $\sigma_F$ are standard deviations of original and reconstructed images, f and F are image contents of pth local window and G is the number of local windows in the image.

The simulation results show that our approach provides a visually appealing output. The proposed DCT based super resolution reconstruction with zonal filter based denoising can reconstruct a super resolution image from a series of blurred, noisy , aliased and down sampled low resolution images, which is demonstrated by two cases as show in Fig. 10 and Fig.11.

In case 1, Fig 10, three LR SAR images of AbuDhabi stadium are considered which are corrupted by Gaussian noise of standard deviation ($\sigma$ =10,15 & 20) and motion blur of angle(10,20 & 30) . Fig 10.(a-c) shows three LR images, Fig10.(d) show the first stage of reconstruction a coarser image , where only DC coefficients are transmitted. Fig. 10(e) second stage of progressive reconstruction consists of DC and first AC coefficients; where as Fig.10 (f-l) shows the different stages of reconstruction. Fig. 10(m) is the super resolution image using our proposed approach. In Fig 11. Low resolution mammogram images are considered and a super resolution mammogram image is reconstructed progressively using our proposed approach in stage X, i.e. fig 11(m).

The results of the proposed SRR are compared with Projection on to Convex Sets (POCS)[17], Papoulis Gerchberg algorithm [19] [20], Iterative Back Propagation (IBP) [18] and FFT based SRR [23] .

**Table I: MSE and PSNR comparison of our proposed approach with other approaches**

| Source Image | POCS | | Paploius Grechberg | | IBP | | FFT SRR | | Proposed SRR | |
|---|---|---|---|---|---|---|---|---|---|---|
| | MSE | PSNR | MSE | PSNR | MSE | PSNR | MSE | PSNR | MSE | PSNR |
| AbudaiSAR | 50.62 | 31.08 | 48.9 | 31.23 | 60.51 | 30.31 | 59.83 | 30.36 | 0.002 | 37.24 |
| Mammogram | 40.63 | 30.06 | 45.1 | 30.04 | 45.78 | 29.95 | 46.08 | 30.86 | 0.035 | 36.87 |
| Baboon | 58.64 | 30.44 | 56.28 | 30.62 | 66.46 | 29.9 | 68.04 | 29.8 | 0.086 | 38.8 |
| lena | 82.015 | 28.99 | 79.52 | 29.12 | 87.21 | 28.72 | 82.48 | 28.96 | 0.061 | 37.95 |

**Table II: ISNR and MSSIM comparison of our proposed approach with other approaches**

| Source Image | POCS | | Paploius Grechberg | | IBP | | FFT SRR | | Proposed SRR | |
|---|---|---|---|---|---|---|---|---|---|---|
| | ISNR | MSSIM | ISNR | MSSIM | ISNR | MSSIM | ISNR | MSSIM | ISNR | MSSIM |
| AbudaiSAR | 3.392 | 0.6783 | 3.96 | 0.625 | 3.35 | 0.761 | 3.39 | 0.68 | 5.49 | 0.833 |
| Mammogram | 3.58 | 0.723 | 4.61 | 0.76 | 3.69 | 0.79 | 3.94 | 0.73 | 5.05 | 0.871 |
| Baboon | 2.8 | 0.671 | 3.09 | 0.663 | 2.13 | 0.711 | 2.51 | 0.702 | 5.87 | 0.778 |
| lena | 3.58 | 0.683 | 3.43 | 0.617 | 3.04 | 0.786 | 4.08 | 0.711 | 5.01 | 0.858 |

## 8. CONCLUSION

In this paper we have proposed a novel DCT based Progressive reconstruction of super resolution image from a set of low resolution images. Our proposed novel zonal filter based denosing retains the low frequency components and discards the high frequency components as most of the noise will be present in the high frequency components. Different zonal masks are applied to obtain optimized results. Experiments are conducted on different natural images corrupted by various noise levels to access the performance of the proposed method in comparison with other methods. The advantage of our proposed approach is that low frequency components are transmitted progressively, at each stage of transmission the quality of the image is enhanced. Finally adaptive interpolation is applied to reconstruct a super resolution image. We are able to obtain very high PSNR and ISNR value when compared to the state of art super resolution reconstruction. We are able to obtain a very high super resolution image with super resolution factor of 2, 4 times than that of the original image. Experimental results show that our proposed method performs quite well in terms of robustness and efficiency.

**Author Biographies**

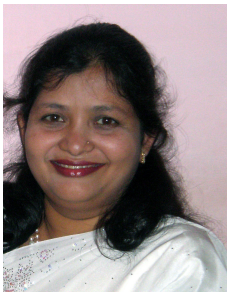

**Ms.Liyakathunisa** completed her Bachelors degree in computer science and Engineering from Gulbarga University and Masters in Software Engineering form Visvesvaraya Technological University. She is presently working at Sri Jayachamarajendra College of Engineering, Mysore in the Department of Computer Science and Engineering. Currently she is pursuing her Ph.D under the guidance of Dr.C.N.Ravi Kumar.  She has published 20 research papers in International Journal / Conferences. Her research interest includes, Medical Image Processing, Satellite Image Processing and Computer Vision.

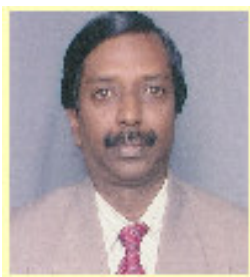

**Dr. C.N. Ravi Kumar** is the founder faculty of the department of Computer Science and Engineering at Sri Jayachamarajendra College of Engineering Mysore.  He is presently working as a Professor and Head of the department. He obtained his Bachelor degree in Electronics and Communication in the year 1979, obtained his Master Degree in the year 1985 and was the first person to obtain the M.Sc. (Engg.) by Research degree from Mysore University. He obtained his doctoral degree in Computer Science and Engineering during the year 2000, under the guidance of Dr. K. Chidananda Gowda, the former Vice-Chancellor of Kuvempu University. He has 105 papers published to his credit in National, International journals and Conferences. He has also been awarded with the Platinum Jubilee Eminent Engineers Award in 2010. His research interest includes Pattern Recognition, Image Processing, Biometrics and Data Mining.